\RequirePackage[svgnames]{xcolor}

\documentclass[11pt,letterpaper]{mystyle}
\usepackage{bxcoloremoji}

\usepackage[all]{hypcap}
\usepackage[svgnames]{xcolor}
\usepackage[comma,authoryear,compress]{natbib}
\usepackage{hyperref}[citecolor=lightblue]

\hypersetup{
    colorlinks = true,
    citecolor = {YaleBlue},
}

\usepackage{algorithm}
\usepackage{algorithmicx}
\usepackage{algpseudocode}
\usepackage{microtype}
\usepackage{graphicx}
\expandafter\def\csname ver@subfig.sty\endcsname{}
\usepackage{booktabs} %
\usepackage{float}
\usepackage{bigstrut}
\usepackage{multirow}
\usepackage{amsmath}
\usepackage{amssymb}
\usepackage{mathtools}
\usepackage{amsthm}
\usepackage{mathrsfs}
\usepackage{nicefrac}
\usepackage{dsfont}
\usepackage{enumitem}
\usepackage{subcaption}
\usepackage{graphicx,subfig}
\usepackage{cleveref}
\usepackage{bxcoloremoji}
\usepackage{float}
\usepackage{fontawesome5}

\usepackage[utf8]{inputenc} %
\usepackage[T1]{fontenc}    %
\usepackage{hyperref}       %
\usepackage{url}            %
\usepackage{booktabs}       %
\usepackage{amsfonts}       %
\usepackage{nicefrac}       %
\usepackage{microtype}      %
\usepackage{graphicx}
\usepackage{subcaption} % Ensure subfigure or subtable environments are used if this package is included.
\usepackage{amssymb}
\usepackage{fdsymbol}
\usepackage{wrapfig}
\usepackage{lipsum}
\usepackage{enumitem}
\usepackage{stackengine}
\usepackage[font=small,labelfont=bf]{caption}
\usepackage{color}
\usepackage{adjustbox}

\usepackage{rotating}
\usepackage{makecell}

\newcommand{\x}{\mathbf{x}}

\definecolor{blanchedalmond}{rgb}{1.0, 0.92, 0.8}
\definecolor{carmine}{rgb}{0.59, 0.0, 0.09}
\definecolor{lightblue}{rgb}{0.22,0.45,0.70}%

\renewcommand{\mathbf}{\boldsymbol}

\makeatletter
\def\Ddots{\mathinner{\mkern1mu\raise\p@
\vbox{\kern7\p@\hbox{.}}\mkern2mu
\raise4\p@\hbox{.}\mkern2mu\raise7\p@\hbox{.}\mkern1mu}}
\makeatother

\definecolor{amaranth}{rgb}{0.9, 0.17, 0.31}
\definecolor{antiquebrass}{rgb}{0.8, 0.58, 0.46}
\definecolor{antiquefuchsia}{rgb}{0.57, 0.36, 0.51}
\definecolor{chromeyellow}{rgb}{0.31, 0.47, 0.26}

\newcommand{\github}{\raisebox{-1.5pt}{\includegraphics[height=1.05em]{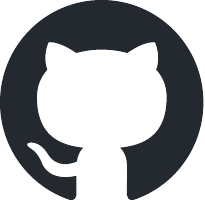}}}

\newcommand{\paperlogo}{\raisebox{-1.5pt}{\includegraphics[height=2.05em]{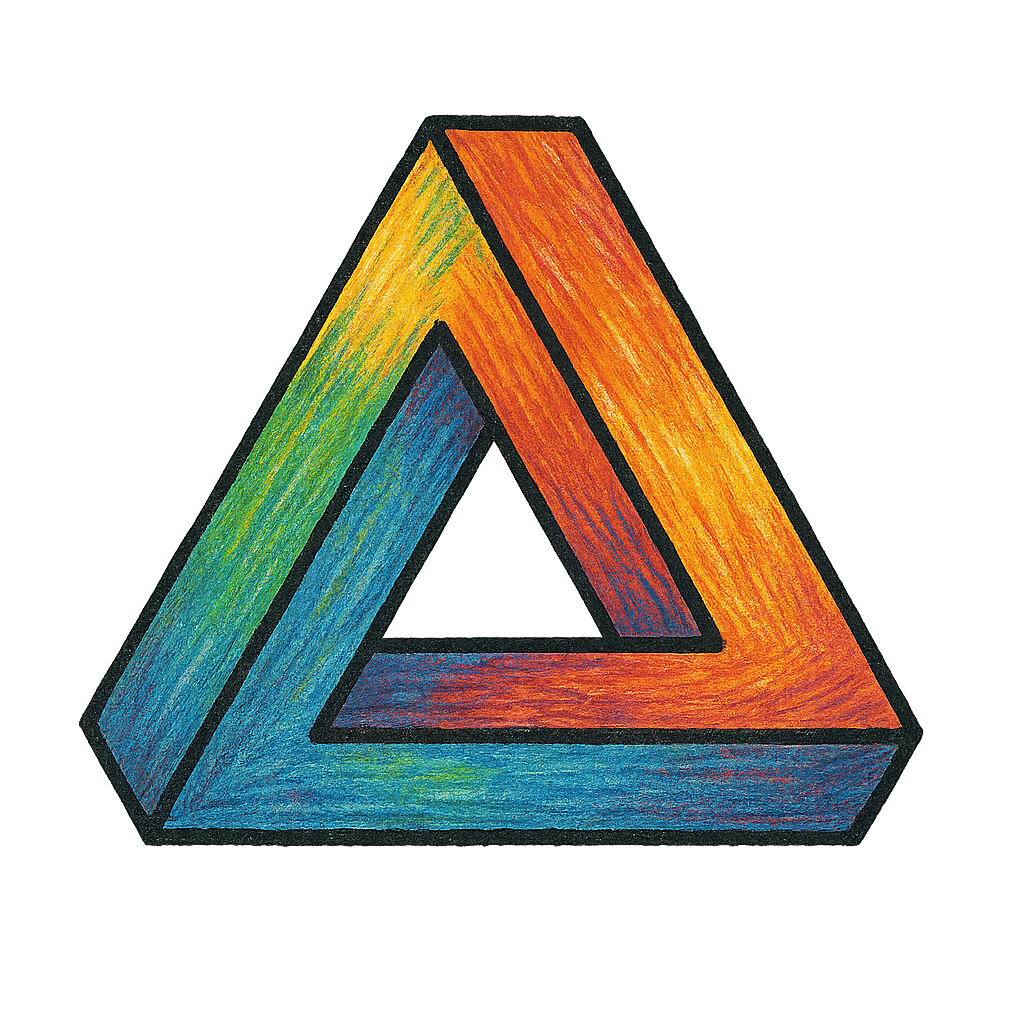}}}

\newtcolorbox{AIbox}[2][]{aibox,title=#2,#1}
\definecolor{lightblue}{rgb}{0.22,0.45,0.70}%
\definecolor{Gray}{gray}{0.95}
\definecolor{Cornsilk}{rgb}{1.0, 0.97, 0.86}

\usepackage{amsmath}

\usepackage[all]{hypcap}

\title{\paperlogo{} HeSRN: Representation Learning On Heterogeneous Graphs via Slot-Aware Retentive Network}

\runningtitle{\paperlogo{} HeSRN: Representation Learning On Heterogeneous Graphs via Slot-Aware Retentive Network}

\author{
  Yifan Lu$^{1}$,
  Ziyun Zou$^{1}$,
  Belal Alsinglawi$^{2,3}$, 
  Islam Al-Qudah$^{4}$, 
  Izzat Alsmadi$^{5}$, \\
  Feilong Tang$^{1,6}$, 
  Pengfei Jiao$^{7}$, 
  Shoaib Jameel$^{8}$,
  Imran Razzak
}

\affil[1]{Mohamed bin Zayed University of Artificial Intelligence, UAE}
\affil[2]{Zayed University, UAE}
\affil[3]{Swinburne University of Technology, Australia}
\affil[4]{Higher Colleges of Technology, UAE}
\affil[5]{Texas A\&M University-San Antonio, United States}
\affil[6]{Monash University, Australia}
\affil[7]{Hangzhou Dianzi University, China}
\affil[8]{University of Southampton, United Kingdom}

\correspondingauthor{Pengfei Jiao; Email: \href{mailto:pjiao@hdu.edu.cn}{pjiao@hdu.edu.cn}}

\begin{document}

\begin{abstract}
Graph Transformers have recently achieved remarkable progress in graph representation learning by capturing long-range dependencies through self-attention. However, their quadratic computational complexity and inability to effectively model heterogeneous semantics severely limit their scalability and generalization on real-world heterogeneous graphs. To address these issues, we propose HeSRN, a novel Heterogeneous Slot-aware Retentive Network for efficient and expressive heterogeneous graph representation learning. HeSRN introduces a slot-aware structure encoder that explicitly disentangles node-type semantics by projecting heterogeneous features into independent slots and aligning their distributions through slot normalization and retention-based fusion, effectively mitigating the semantic entanglement caused by forced feature-space unification in previous Transformer-based models. Furthermore, we replace the self-attention mechanism with a retention-based encoder, which models structural and contextual dependencies in linear time complexity while maintaining strong expressive power. A heterogeneous retentive encoder is further employed to jointly capture both local structural signals and global heterogeneous semantics through multi-scale retention layers. Extensive experiments on four real-world heterogeneous graph datasets demonstrate that HeSRN consistently outperforms state-of-the-art heterogeneous graph neural networks and Graph Transformer baselines on node classification tasks, achieving superior accuracy with significantly lower computational complexity.

\vspace{5mm}


\vspace{5mm}

% \coloremojicode{1F4C5} \textbf{Date}: April 24, 2025

% \coloremojicode{1F3E0} \textbf{Projects}: \href{https://wangrongsheng.github.io}{https://wangrongsheng.github.io}

\github{} \textbf{Code Repository}: \href{https://github.com/csyifan/HeSRN}{https://github.com/csyifan/HeSRN}

\coloremojicode{1F4E7} \textbf{Contact}: \href{mailto:Yifan.Lu@mbzuai.ac.ae}{Yifan.Lu@mbzuai.ac.ae}

\end{abstract}

\maketitle

\newpage

\section{Introduction}
Graph transformers have emerged as a novel architecture for modeling graph-structured data and have been widely applied in graph representation learning tasks, such as node classification~\cite{chenleveraging}, link prediction~\cite{shomer2024lpformer}, and recommendation systems~\cite{chen2024sigformer}. Unlike traditional graph neural network (GNN) methods~\cite{das2024ags,luan2024graph, zou2025loha}, which implement neighborhood message passing to model local relations in graph data, graph transformer leverages a global attention mechanism~\cite{velivckovic2018graph,brodyattentive} to capture long-range dependencies within the graph, offering a more superior capability in deep modeling.

In conventional GNNs, node representations are typically updated at each layer by aggregating information from neighboring nodes, relying on local neighborhood message passing~\cite{ji2024regcl}. As the number of layers increases, dependencies between nodes can be progressively propagated, but long-range dependencies remain constrained by the adjacency structure~\cite{wang2024graph}. This limitation arises because each layer of a GNN aggregates and weights the representations of a node and its neighbors, causing the similarity of node representations to gradually increase~\cite{shen2024resisting}. Furthermore, when nodes need to aggregate information from higher-order neighbors, the information passed through multiple layers tends to be overly compressed in the fixed representation space~\cite{huanguniversal}, leading to information loss. In contrast, graph transformers model dependencies between nodes using a global self-attention mechanism, which is not limited by the adjacency structure. Each node can attend to all other nodes with varying attention weights, enabling stronger expressive power~\cite{zhu2023structural}.

However, due to the need for computing global self-attention~\cite{rampavsek2022recipe,xingless}, the complexity of the self-attention mechanism in graph transformers is typically \( O(N^2) \). In graph data, where the number of nodes is usually very large, this results in significant computational and storage overheads~\cite{wu2024simplifying,kong2023goat}. Additionally, since graph transformer requires maintaining the Key-Value cache~\cite{zhang2024cached,brandon2024reducing} for historical states during forward propagation, it increases latency in the propagation process. To reduce the complexity of the self-attention mechanism, NodeFormer~\cite{wu2022nodeformer} samples important neighbors of the target nodes and sparsifies the attention matrix. However, this approach sacrifices long-range dependencies and lacks stability when capturing global information.

Another challenge arises from the nature of graph data itself. Real-world graph data is often heterogeneous~\cite{zhang2019heterogeneous,yang2023simple,wang2022survey}, consisting of different types of nodes and edges, which form the complex semantics of heterogeneous graphs, such as chemical molecular graphs~\cite{wu2023molformer}, citation graphs~\cite{yang2023revisiting}, and knowledge graphs~\cite{ruiz2024high}. In molecular graphs, different types of atoms (e.g., $N, H, O$) form nodes in the graph, while the chemical bonds between atoms (e.g., $N$-$O$, $N$-$H$, $H$-$O$) form different types of edges in the heterogeneous graph. In this context, specific substructures like benzene rings and carboxyl groups, when combined, express higher-level heterogeneous semantics. Since graph transformers are designed for homogeneous graphs, they lack the ability to model heterogeneous semantics, making it difficult to learn the complex semantics within heterogeneous graphs. Recently, some heterogeneous graph transformer architectures have been proposed for heterogeneous graph representation learning. HINormer~\cite{mao2023hinormer} uses local structure encoders and relationship encoders to capture local heterogeneous node representations but ignores higher-order semantics of heterogeneous graphs. Meanwhile, during initialization, HINormer uses a predefined GCN~\cite{kipf2016semi} function to forcibly aggregate all types of nodes into a single feature space. This causes features of different fine-grained types to become entangled at initialization, and their semantics cannot be properly aligned. PHGT~\cite{lu2024heterogeneous} builds on HINormer by incorporating prior knowledge of heterogeneous meta-paths and adds sampled cluster information into tokens to represent global semantics. However, this approach further increases the model's complexity and limits its application in real-world scenarios. At the same time, existing methods are still constrained by the complexity of the Transformer architecture, and the development of next-generation architectures~\cite{retnet} has received much attention.

To address the aforementioned challenges, we propose HeSRN, a novel Heterogeneous Slot-aware Retentive Network for representation learning on heterogeneous graphs. HeSRN introduces a slot-aware structure encoder that explicitly disentangles node-type semantics by projecting heterogeneous node features into independent slots and aligning their distributions through slot alignment and retention-based fusion. To further capture both structural and contextual dependencies with high efficiency, HeSRN replaces self-attention with a retention mechanism, enabling linear-time sequence modeling while maintaining strong expressive capacity. A heterogeneous retentive encoder is then employed to jointly model local structural signals and global heterogeneous semantics through multi-scale retention layers, achieving efficient and scalable graph representation learning.

In summary, our main contributions are as follows:
\begin{itemize}
\item We propose HeSRN, a new heterogeneous graph learning framework that unifies slot-aware semantic modeling and retentive sequence encoding, effectively reducing the complexity and memory cost of Transformer architectures while improving expressive power.
\item We design a heterogeneous retention mechanism that integrates type-aware semantic fusion and global contextual modeling, enabling efficient and expressive representation learning across heterogeneous node types.
\item Extensive experiments on four real-world heterogeneous graph datasets demonstrate that HeSRN consistently outperforms state-of-the-art heterogeneous graph neural networks and graph transformer models on node classification tasks.
\end{itemize}

\section{Related Work}
\subsection{Heterogeneous Graph Neural Networks}
Heterogeneous Graph Neural Networks (HGNNs) have attracted significant attention in recent years due to their ability to model multiple types of nodes and edges, thereby capturing complex heterogeneous semantics. Existing HGNN approaches can be broadly divided into two categories: meta-path-based methods and path-free methods. Meta-path-based models require predefined meta-paths to characterize semantic connections among different node types and capture heterogeneous relations through meta-path-based sampling and aggregation. For instance, the Heterogeneous Graph Attention Network~\cite{wang2019heterogeneous} employs a dual-level attention mechanism, where node-level attention learns the importance of intra-type neighbors and semantic-level attention adaptively integrates multiple meta-path semantics. MAGNN~\cite{fu2020magnn} further extends this idea by incorporating intermediate nodes along each meta-path, enabling finer-grained contextual aggregation of semantic dependencies. GTN~\cite{yun2019graph} eliminates the need for manually defined meta-paths by introducing learnable soft selection matrices that automatically construct latent meta-paths in an end-to-end manner. In contrast, path-free HGNNs dispense with manually designed meta-paths and instead learn heterogeneous semantics directly through neighborhood aggregation. Representative methods include RGCN~\cite{schlichtkrull2018modeling}, which designs relation-specific transformation matrices to perform relational convolutions; Simple-HGN~\cite{lv2021we}, which simplifies the model architecture by removing complex attention mechanisms while maintaining efficiency through relation encoding and residual aggregation; and HetGNN~\cite{zhang2019heterogeneous}, which encodes multimodal content features for each node type and performs type-level aggregation for joint representation learning. With the deepening understanding of HGNN architectures, recent research has extended toward more complex and practical directions, such as molecular graph modeling in drug discovery~\cite{ni2025robust}, interpretability of heterogeneous representations~\cite{li2023heterogeneous}, and the integration of HGNNs with Large Language Models~\cite{tang2024higpt}, which together are driving the advancement of interpretable and multimodal graph intelligence. However, these methods are still constrained by the message passing mechanism, which weakens their performance in graph representation learning tasks. 

\subsection{Graph Transformers}
Graph Transformers (GTs) have recently emerged as a powerful architecture for graph representation learning, achieving remarkable performance across a variety of graph-based tasks. Depending on the graph type they target, GTs can be broadly categorized into homogeneous and heterogeneous graph transformers. For homogeneous graphs, ANS-GT~\cite{zhang2022hierarchical} integrates adaptive neighborhood sampling with attention mechanisms to efficiently handle large-scale graphs. NodeFormer~\cite{wu2022nodeformer} leverages kernelized attention to approximate the softmax attention and thus reduces computational cost. NAGphormer~\cite{chen2022nagphormer} models both node- and graph-level dependencies through neighborhood aggregation and global contextual learning, enabling effective multi-scale representation. Building upon homogeneous architectures, heterogeneous graph transformers incorporate semantic-level modeling to capture the diversity of node and edge types. HINormer~\cite{mao2023hinormer} introduces normalization and semantic encoding modules to effectively learn heterogeneous semantic dependencies. PHGT~\cite{lu2024heterogeneous} fuses pre-learned meta-path–based semantic and structural encodings to obtain fine-grained heterogeneous node representations. However, despite their success, these approaches still operate within the standard Transformer framework and thus suffer from the quadratic complexity of the attention mechanism. To address this limitation, HeSRN adopts a retentive architecture that eliminates the need to maintain explicit attention matrices, substantially reducing computational and memory overhead while preserving modeling capacity.

\begin{figure}[!ht]
  \centering
  \includegraphics[width=\textwidth]{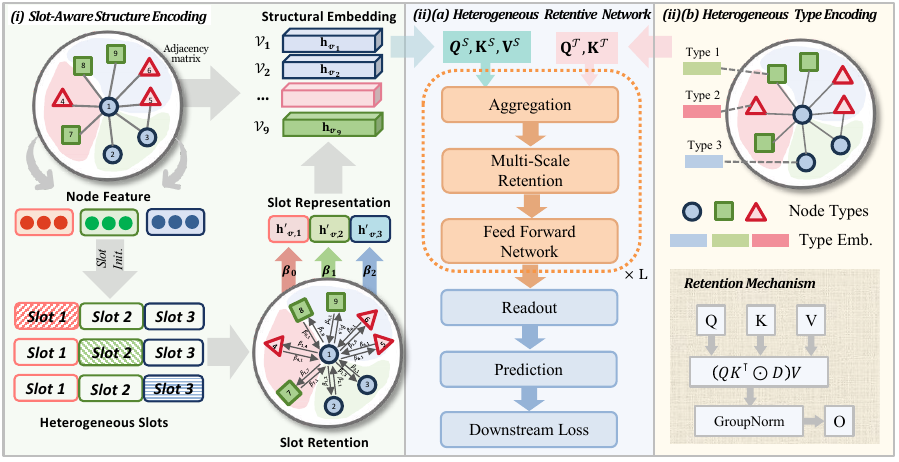}
  \caption{The architecture of our proposed HeSRN. First, the slot-aware structure encoding module assigns a slot to each node according to its heterogeneous type and employs slot retention to learn the importance of different slots, aggregating them into a slot representation. Then, the slot representations are combined with the heterogeneous graph structure to obtain the structural representations. Subsequently, these structural representations are fed into the retentive network, where they are integrated with heterogeneous type embeddings to perform multi-scale retention computations. After passing through $L$ retentive layers, the model learns the node representations used for downstream tasks.}
\label{model}
\end{figure}

\section{Notations and Preliminaries}

In this section, we introduce the notations and fundamental concepts used throughout this paper, including the definition of heterogeneous graphs and the retentive architecture that serves as the foundation of our proposed model.

\subsection{Heterogeneous Graph}
A heterogeneous graph \( \mathcal{G} = (\mathcal{V}, \mathcal{E}, \mathcal{T}_\mathcal{V}, \mathcal{T}_\mathcal{E}) \) consists of a set of nodes \( \mathcal{V} \) and edges \( \mathcal{E} \), where each node \( v \in \mathcal{V} \) is associated with a type \( \phi(v) \in \mathcal{T}_\mathcal{V} \) and each edge \( e \in \mathcal{E} \) is associated with a type \( \psi(e) \in \mathcal{T}_\mathcal{E} \). $\phi(\cdot)$ and $\psi(\cdot)$ are mapping functions of node type and edge type. The adjacency matrix of $\mathcal{G}$ is denoted as \( \mathbf{A} \in \mathbb{R}^{N \times N} \), where \( \mathbf{A}_{ij}=1 \) when $e_{ij}$ exists and $N$ is the number of nodes. The feature \( \mathbf{X} \) of $\mathcal{G}$ is a collection of feature matrices \( \{\mathbf{X}_t\}^{|\mathcal{T}_\mathcal{V}|}_{t=0} \), and different node types have different feature dimensions. 

\subsection{Retentive Architecture}

A typical Retentive layer~\cite{retnet} is constructed from two essential submodules: a \textit{multi-scale retention module} (MSR) and a \textit{feed-forward network} (FFN). For conciseness, we describe here the single-head variant of the MSR.

Let the input sequence be denoted as:
$\tilde{\mathbf{H}} = [\tilde{\mathbf{h}}_1, \tilde{\mathbf{h}}_2, ..., \tilde{\mathbf{h}}_n]^{\mathrm{T}} \in \mathbb{R}^{N \times d}$,
where $d$ is the hidden dimensionality and $\tilde{\mathbf{h}}_i \in \mathbb{R}^{d}$ represents the hidden state at step $i$.
The MSR projects the input into query, key, and value spaces through three parameter matrices:
$\mathbf{W}_Q, \mathbf{W}_K, \mathbf{W}_V \in \mathbb{R}^{d \times d}$, given by:
\begin{gather}
    \mathbf{Q}=(\tilde{\mathbf{H}}\mathbf{W}_Q)\odot\Theta,\quad
    \mathbf{K}=(\tilde{\mathbf{H}}\mathbf{W}_K)\odot\bar{\Theta},\quad
    \mathbf{V}=\tilde{\mathbf{H}}\mathbf{W}_V, \label{qkv_modified} \\
    \Theta_n=e^{in\theta},\quad 
    D_{nm}=\left\{
        \begin{array}{ll}
            \gamma^{n-m}, & \quad n\geq m\\[3pt]
            0, & \quad n<m
        \end{array}
    \right. \\[3pt]
    \mathrm{Retention}(\tilde{\mathbf{H}})=
    \big((\mathbf{Q}{\mathbf{K}}^{\top})\odot D\big)\mathbf{V}.
\end{gather}
In this formulation, $\Theta$ and its conjugate $\bar{\Theta}$ introduce complex-valued positional modulation, while $D$ represents a lower-triangular decay mask whose entries exponentially diminish with distance, controlled by a learnable factor $\gamma \in (0,1)$.
The element-wise operation $\odot$ jointly applies phase modulation and exponential attenuation, providing an efficient linear-time approximation to long-range temporal interactions.

The MSR output is then fed into a two-layer FFN module equipped with Layer Normalization (LN)~\cite{lei2016layer} and residual connections~\cite{he2016deep}.
The output of the $l$-th Retentive layer is formulated as:
\begin{equation}
    \hat{\mathbf{H}}^{(l)} = \mathrm{LN}\big(\mathrm{MSR}(\tilde{\mathbf{H}}^{(l-1)}) + \tilde{\mathbf{H}}^{(l-1)}\big),
    \label{eq:msr_ln}
\end{equation}
\begin{equation}
    \tilde{\mathbf{H}}^{(l)} = \mathrm{LN}\big(\mathrm{FFN}(\hat{\mathbf{H}}^{(l)}) + \hat{\mathbf{H}}^{(l)}\big).
    \label{eq:ffn_ln}
\end{equation}

By stacking multiple Retentive layers, the network progressively aggregates temporal information through phase-modulated and exponentially decayed memory states.
The final representation $\tilde{\mathbf{H}}^{(L)} \in \mathbb{R}^{N \times d}$ serves as the contextual embedding for subsequent downstream tasks.

\section{Methodology}
In this section, we formally present HeSRN. Figure~\ref{model} illustrates the overall architecture of HeSRN. HeSRN consists of two main components: a slot-aware structure encoding module and a retentive network. The slot-aware structure encoding module first assigns a distinct slot to each node based on its heterogeneous type. Through slot retention, the model learns the relative importance of different slots and aggregates them into comprehensive slot representations. These representations are then combined with the heterogeneous graph structure to produce structural representations that capture both local and type-specific information. Next, the structural representations are fed into the retentive network, where they are fused with heterogeneous node embeddings to perform multi-scale retention operations. This process allows the model to capture dependencies across various semantic and structural levels. After passing through $L$ retentive layers, HeSRN produces refined node representations that can be directly utilized for downstream tasks.

\subsection{Slot-Aware Structure Encoding}
Traditional Graph Transformer architectures generate node sequences by inputting the entire graph and then compute the self-attention among the nodes in the sequence. This approach has two main limitations: 1) The self-attention mechanism operates under the assumption that all nodes reside in a shared feature space, which naturally holds true for homogeneous graphs. However, in heterogeneous graphs, nodes of different types carry distinct semantics and exist in separate feature spaces. 2) The self-attention mechanism only calculates pairwise interactions within the receptive field, overlooking the explicit neighborhood connections in the graph. It has been demonstrated that incorporating graph structural information can enhance the expressive power and generalization ability of the network~\cite{lu2023neighborhood,zhu2023structural,boker2024fine}. Although HINormer~\cite{mao2023hinormer} and PHGT~\cite{lu2024heterogeneous} incorporate structural encodings, they simply treat the graph as homogeneous and use GCN to aggregate initial node representations. The heterogeneous information is only introduced during the Transformer computation. This initialization path forces nodes of different types to aggregate together, causing their representations to become entangled and thereby weakening the model’s performance within the Transformer. We introduce heterogeneous slots to independently learn the type information of each node, and employ retention mechanism to capture the importance of different types, thereby integrating heterogeneous semantics into the structural encoding.

\subsubsection{Slot Preparation}
Since different types of node features exist in spaces of varying dimensions, a mapping function is first used to project the features of different node types into a unified feature dimension $d$:
\begin{equation}
    \mathbf{H}_t=\mathbf{X}_t\mathbf{W}_t,
\end{equation}
where $\mathbf{W}_t \in \mathbb{R}^{d_t \times d}$. $\mathbf{H}_t \in \mathbb{R}^{\mathcal{V}_t \times d}$ is the feature representation matrix of type-$t$ nodes. Then, all nodes are grouped into separate slots according to their types, resulting in a slot matrix $\mathcal{H} \in \mathbb{R}^{N\times |\mathcal{T}_\mathcal{V}| \times d}$:
\begin{equation}
\mathcal{H}_{v,s,:}=\begin{cases}\mathbf{h}_{v},&\mathrm{if}\phi(v)=s,\\\mathbf{0}_{D},&\mathrm{otherwise},\end{cases}
\end{equation}
where $\mathbf{h}_{v}$ is the $d$-dimensional embedding of node $v$ obtained from $\mathbf{H}_{\phi(v)}$.

\subsubsection{Slot Alignment}
Although the slot features of different types obtained in the previous step have been mapped to the same embedding dimension $d$, there still exist significant distributional differences among the slots. Therefore, we need to further align and normalize the semantic spaces of all slots. The process is as follows:
\begin{equation}
\tilde{\mathcal{H}}_{v,s,:}=\phi_{slot}\left(\mathrm{LN}\left(\mathcal{H}_{v,s,:}U_{s}+b_{s}\right)\right),
\label{slotalign}
\end{equation}
where $U_{s}\in \mathbb{R}^{d \times d}$ is the learnable weights. The slot alignment serves three main purposes: Firstly, by introducing nonlinear transformations, it enables the model to capture more complex feature relationships, enhancing the representational capacity within each slot. Secondly, it assigns independent linear mapping weights to each slot, allowing for feature alignment and semantic normalization within the slots. Thirdly, the LayerNorm operation balances the scale of features across slots, helping to ensure consistent distributions during subsequent computations.

\subsubsection{Slot Retention}
After slot alignment, each node $v$ is represented by a slot tensor $\tilde{\mathcal{H}}_{v,:,:}\in\mathbb{R}^{1\times|\mathcal{T}_\mathcal{V}|\times d}$. To model the semantic dependencies among heterogeneous slots while maintaining linear complexity, we employ a \textit{retention mechanism} to replace the standard self-attention used in previous Graph Transformers. Specifically, we project the aligned slot representations into query, key, and value spaces:
\begin{equation}
\mathbf{Q}_{v,s} = \tilde{\mathcal{H}}_{v,s,:}\mathbf{W}_s^Q, \quad
\mathbf{K}_{v,s} = \tilde{\mathcal{H}}_{v,s,:}\mathbf{W}_s^K, \quad
\mathbf{V}_{v,s} = \tilde{\mathcal{H}}_{v,s,:}\mathbf{W}_s^V,
\end{equation}
where $\mathbf{W}_s^Q, \mathbf{W}_s^K, \mathbf{W}_s^V \in \mathbb{R}^{d\times d}$ are learnable parameters shared across all slots.

To model the relative importance between slots, we introduce an exponentially decaying lower-triangular kernel matrix $\mathbf{D}\in\mathbb{R}^{C\times C}$ defined as
\begin{equation}
\mathbf{D}_{s_1,s_2}=
\begin{cases}
\gamma^{\,s_1-s_2}, & s_1\ge s_2,\\
0, & s_1 < s_2,
\end{cases}
\end{equation}
where $\gamma\in(0,1)$ is a learnable decay coefficient that controls the retention strength. The retention operation is then formulated as:
\begin{equation}
\mathrm{Retention}(\tilde{\mathcal{H}}_{v,:,:}) 
= \big((\mathbf{Q}_v\mathbf{K}_v^{\top})\odot\mathbf{D}\big)\mathbf{V}_v,
\label{eq:slot_retention}
\end{equation}

To stabilize training and enhance representation expressiveness, we apply a residual connection and a learnable scaling factor:
\begin{equation}
\mathcal{H}'_{v,:,:} = \tilde{\mathcal{H}}_{v,:,:} + \alpha_s \cdot
\mathrm{Retention}(\tilde{\mathcal{H}}_{v,:,:}),
\end{equation}
where $\alpha_s$ is a learnable parameter controlling the fusion intensity.
The resulting $\mathcal{H}'_{v,:,:}$ captures the heterogeneous correlations among different node-type slots while preserving type-specific semantics.

Finally, all slot representations of node $v$ are fused through a semantic-level attention mechanism to obtain a unified node embedding $\mathbf{h}'_v \in \mathbb{R}^{1\times d}$ that integrates multi-type information in a slot-aware manner:

\begin{gather}
\tilde{\mathbf{s}}_{v,s} = \tanh(\mathbf{W}^{s}_a\,\mathbf{h}'_{v,s} + \mathbf{b}^{s}_a),
\\
\tilde{\beta}_{v,s} = \mathbf{p}^{\top}\tilde{\mathbf{s}}_{v,s},
\\
\beta_{v,s} = 
\frac{\exp(\tilde{\beta}_{v,s})}
     {\sum_{s'=1}^{|\mathcal{T}_\mathcal{V}|}\exp(\tilde{\beta}_{v,s'})},\quad
\mathbf{h}'_v = 
\sum_{s=1}^{|\mathcal{T}_\mathcal{V}|}\beta_{v,s}\,\mathbf{h}'_{v,s}, 
\end{gather}
where $\mathbf{W}^{s}_a\in\mathbb{R}^{d\times a}$ and $\mathbf{b}^{s}_a\in\mathbb{R}^{a}$ are learnable parameters that project slot features into an attention space of dimension $a$, $\mathbf{p}\in\mathbb{R}^{a}$ is a global semantic query vector, and $\beta_{v,s}$ represents the semantic importance of slot $s$ for node $v$. 

\subsubsection{Structure Encoding with Slot}
To better integrate the learned slot-aware node representations with the graph structure, we apply an aggregation function that combines the adjacency matrix \( \mathbf{A} \) with the current node representations to obtain the updated embeddings:
\begin{equation}
    \mathbf{h}^{(l)}_{v}=\mathrm{Aggregation}\left(\mathbf{h'}_v^{(l-1)},\left\{\mathbf{h'}_j^{(l-1)}\mid j\in\mathcal{N}_v\right\}\right),
\end{equation}
where $\mathcal{N}_v$ represents the neighbor of node $v$ and $\mathbf{h'}^{(0)}_{v} = \mathbf{h}_{v}'$.

 % Compared to Transformer, it substitutes the self-attention mechanism with a retention mechanism, reducing the high memory overhead associated with self-attention calculations.

\subsection{Heterogeneous Graph Retentive Network}
The retentive network was first introduced in RetNet~\cite{retnet} as a network layer for sequence modeling, primarily used in language models~\cite{huang2024leret,guo2024vl}. By combining the advantages of both Recurrent Neural Network~\cite{zaremba2014recurrent} and Transformer~\cite{vaswani2017attention} architectures, it achieves parallel training, low inference costs, and competitive performance. Inspired by RetNet, we design the heterogeneous graph retentive network, which not only reduces the computational memory requirements for heterogeneous graph representation learning but also explicitly integrates heterogeneous semantic learning with node representation within the retentive network framework. We define the target node token sequence $\mathbf{H}_{v}^{S}=[ \mathbf{h}^{(l)}_{v},\mathbf{h}^{(l)}_{1},\mathbf{h}^{(l)}_{2},...,\mathbf{h}^{(l)}_{l-1}]$ as the concatenation of the target node $v$ and all its neighbors, which serves as the input to the Retentive Network. The entire input of all target nodes is \( \mathbf{H}^{S} \). Our experiments reveal that performance does not always improve with an increasing number of neighborhood nodes. Typically, when the sequence length $l$ is set to a moderate value, i.e. $l=50$, the model achieves optimal performance.

\subsubsection{Retention} 
Given the node token sequence \( \mathbf{H}^{S} \), similar to slot retention, we obtain the query matrix \( \mathbf{Q}^{S} \), key matrix \( \mathbf{K}^{S} \), and value matrix \( \mathbf{V}^{S} \) by mapping transformation, which contains learnable weights \( \mathbf{W} \). The graph retention layer can be obtained in the following form:
\begin{gather}
    \mathbf{Q}^{S}=(\mathbf{H}^{S}\mathbf{W}^{S}_Q)\odot\Theta,\quad\mathbf{K}^{S}=(\mathbf{H}^{S}\mathbf{W}^{S}_K)\odot\bar{\Theta},\quad\mathbf{V}^{S}=\mathbf{H}^{S}\mathbf{W}^{S}_V, \label{qkv} \\
    \Theta_n=e^{in\theta},\quad D_{nm}=\left\{\begin{array}{ll}\gamma^{n-m},&\quad n\geq m\\0,&\quad n<m\end{array}\right. \\
    \mathrm{Retention}(\mathbf{\mathbf{H}^{S}})=(\mathbf{(Q^{S}K^{S}})^\top\odot D)\mathbf{V}^{S},
\end{gather}
where $\mathbf{W}^{S}_Q$, $\mathbf{W}^{S}_K$, $\mathbf{W}^{S}_V$ are learnable weights. $D$ is a causal mask and exponential decay matrix used to control the weakening of the correlation between different positions in the node token sequence as the distance increases. In the computation of the retention, an important aspect is the relative position encoding. We provide a detailed description of the relative position encoding operation in HeSRN.

\subsubsection{Relative Position Encoding}
Relative position encoding is used to help the model understand the relative positional relations between elements in a sequence. In sequence models, such as Transformers, there is no inherent mechanism to perceive the positional relationships of input elements due to their attention-based mechanism. Therefore, various position encoding methods~\cite{kazemnejad2024impact,ruoss2023randomized,wu2021rethinking} have been proposed to encode positional information. In HeSRN, we adopt \( xPos \)~\cite{sun2023length} relative position encoding to encode relative positions. \( xPos \) applies a rotation matrix to each position, where the rotation matrix depends on the position itself and a learnable scaling factor, enabling \( xPos \) to encode positional information in a continuous and smooth manner across different scales. In heterogeneous graph representation learning, \( xPos \) can capture graph information at multiple scales. For example, in chemical molecular graphs~\cite{jiang2023pharmacophoric}, atomic interactions can manifest at various scales, from local chemical bonds (short-range interactions) to broader molecular conformations (long-range interactions). Flexibly adjusting the scale of \( xPos \) position encoding helps the model effectively capture these multi-scale features. We use \( n \) and \( m \) to represent position indices, and then obtain the query matrix \( \mathbf{Q} \) and key matrix \( \mathbf{K} \) from Equation~\ref{qkv}.
\begin{gather}
    \mathbf{q}_{i,n}=(\mathbf{H}_{i,n}^S\mathbf{W}_Q)\cdot e^{in\theta}\cdot\gamma^n=(\mathbf{H}_{i,n}^S\mathbf{W}_Q)\cdot (\gamma e^{i\theta})^n, \\
    \mathbf{k}_{i,m}=(\mathbf{H}_{i,m}^S\mathbf{W}_K)\cdot e^{i(-m)\theta}\cdot\gamma^{(-m)} = (\mathbf{H}_{i,m}^S\mathbf{W}_K)\cdot(\gamma e^{i\theta})^{-m},
\end{gather}
where $\gamma$ is the decay factor in the $D$ matrix, which is explicitly incorporated into the $\mathbf{Q}$, $\mathbf{K}$ calculations. \((\gamma e^{i\theta})^n\) and \((\gamma e^{i\theta})^{-m}\) represent the implementation of the \( xPos \) relative position encoding in retention, directly reflecting the relative distance between two positions. \( xPos \) can be defined as: 
\begin{equation}
xPos(n,m) = (\gamma e^{i\theta})^{n-m}.
\end{equation}
The correlation decay is controlled by \( \gamma \). The relative position encoding \( xPos(n,m) \) not only captures the relative positional relations between elements but also simulates the diminishing correlation as the relative distance increases, which is consistent with the use of \( (\gamma e^{i\theta})^n \) and \( (\gamma e^{i\theta})^{-m} \) in the query matrix \( \mathbf{Q} \) and key matrix \( \mathbf{K} \).

\subsubsection{Heterogeneous Retention Encoder}
Heterogeneous graphs possess rich heterogeneity, where different types of semantics naturally interact within the graph. To model the both local and global heterogeneity in heterogeneous graphs, we propose a novel heterogeneous retention encoder (HRE) to learn the heterogeneous semantic. Specifically, we encode local heterogeneity through a type neighborhood aggregation method and explicitly interact it with the heterogeneous graph structure to obtain type embeddings. In particular, the type embeddings are combined with node embedding $\mathbf{h}^{(l)}_{v}$ and processed through the HRE to produce type-aware global retention. 

For the heterogeneous graph \( \mathcal{G} \), we encode the node type \( \mathcal{T}_\mathcal{V} \) as a one-hot vector to obtain the type matrix \( \mathbf{T}\in\mathbb{R}^{|\mathcal{T}_\mathcal{V}|\times|\mathcal{T}_\mathcal{V}|} \). Then, to combine the type information with the structural information, we efficiently fuse them through a simple structural aggregation function. Since the number of node types \( |\mathcal{T}_\mathcal{V}| \) is typically a small value, this aggregation step does not introduce significant computational overhead. The type embedding $\mathbf{h}^{\mathcal{T}(l)}_{i}$ of node \( i \) can be obtained as follows:
\begin{equation}
    \mathbf{h}^{\mathcal{T}(l_t)}_{i}=\mathrm{Aggregation}\left(\mathbf{h}^{\mathcal{T}(l_t-1)}_{i},\left\{\mathbf{h}^{\mathcal{T}(l_t-1)}_{j}\mid j\in\mathcal{N}_i\right\}\right),
\end{equation}
where $\mathbf{h}^{\mathcal{T}(0)}_{i} = \mathrm{T}[\phi(i),:]$. In practice, we use simple GCN~\cite{kipf2016semi} as the aggregation function. After \( l_t \) layers of aggregation, we obtain the type embeddings $\mathbf{H}^{\mathcal{T}} \in \mathbb{R}^{N\times|\mathcal{T}_\mathcal{V}|}$ that incorporate structural information. We can then use the type embeddings for the type retention calculation to adjust the global feature representations and capture semantic interactions between nodes. Therefore, the heterogeneous retention can be obtained as follows:
\begin{gather}
    \mathbf{Q}^{\mathcal{T}}=(\mathbf{H}^{\mathcal{T}}\mathbf{W}^{\mathcal{T}}_Q)\odot\Theta,\quad\mathbf{K}^{\mathcal{T}}=(\mathbf{H}^{\mathcal{T}}\mathbf{W}^{\mathcal{T}}_K)\odot\bar{\Theta}, \\
    \mathrm{He\text{-}Retention}(\mathbf{\mathbf{H}^{S}},\mathbf{H}^{\mathcal{T}})=((\mathbf{Q}^{S}(\mathbf{K}^{S})^\top+\beta_\mathcal{T}\cdot\mathbf{Q}^{\mathcal{T}}(\mathbf{K}^{\mathcal{T}})^\top) \odot D)\mathbf{V}^{S},
\label{herp}
\end{gather}
where $\mathbf{W}^{\mathcal{T}}_Q$, $\mathbf{W}^{\mathcal{T}}_K \in \mathbb{R}^{|\mathcal{T}_\mathcal{V}|\times|\mathcal{T}_\mathcal{V}|}$ are learnable weights, and $\beta_\mathcal{T}$ is the hyperparameter to modulate the weight of heterogeneous semantics. Finally, the heterogeneous retention is extended to multi-head.

\subsubsection{Multi-scale retention}
To enable the model to learn information at different scales, we introduce multi-scale retention (MSR). In each layer, we use \( h \) retention heads, with each head assigned a different \( \gamma \) value. The \( \gamma \) values remain constant across different layers. Additionally, we use a skew gate to enhance the non-linearity of the retention layer. At the same time, we employ a swish gate to further enhance the non-linearity of the retention layers. The multi-scale retention layer is represented as follows:
\begin{gather}
\gamma=1-e^{\mathrm{linspace}(\log1/32,\log1/512,h)},\\
\mathrm{head}_{i}=\mathrm{He\text{-}Retention}(\mathbf{\mathbf{H}^{S}},\mathbf{H}^{\mathcal{T}}), \text{with}\ \gamma_{i},\\
\mathbf{Y}=\mathrm{GroupNorm}_{h}(\boldsymbol{\bigcup}_{i=0}^{h}(\mathrm{head}_{1},\cdots,\mathrm{head}_{h})),\\
\mathrm{MSR}(\mathbf{\mathbf{H}^{S}},\mathbf{H}^{\mathcal{T}})=(\mathrm{swish}(\mathbf{\mathbf{\mathbf{H}^{S}}W_{G}})\odot\mathbf{Y})\mathbf{W_{O}},
\end{gather}
where $\mathbf{W_{G}}$, $\mathbf{W_{O}} \in \mathbb{R}^{N\times d}$ are learnable parameters, and GroupNorm~\cite{wu2018group} normalizes the output of each head.

\subsubsection{Overall HeSRN layer}
For the \( l \)-th layer of the HeSRN, we stack multi-scale retention (MSR) and the feed-forward network (FFN) to construct the model. The FFN consists of two linear transformations that capture the interactions between different latent dimensions, followed by a GeLU~\cite{hendrycks2016gaussian} activation:
\begin{equation}
\mathrm{FFN}(\mathbf{X}) = \mathrm{GeLU}(\mathbf{X}\mathbf{W}^1_f)\mathbf{W}^2_f,    
\end{equation}
where $\mathbf{X}$ is the input, $\mathbf{W}^1_f$ and $\mathbf{W}^2_f$ are learnable weights.
The overall HeSRN layer can be expressed as follows:
\begin{gather}
\mathbf{Y}^{(l-1)} = \mathrm{MSR}(\mathrm{LN}(\mathbf{H}^{S(l-1)}),\mathrm{LN}(\mathbf{H}^{\mathcal{T}(l-1)}))+\mathbf{H}^{S(l-1)},\\
\mathbf{H}^{S{(l)}} = \mathrm{FFN}(\mathrm{LN}(\mathbf{Y}^{(l-1)})+\mathbf{Y}^{(l-1)},
\end{gather}

\begin{table}[t]
\centering
\caption{Summary of datasets used in the study.}
\renewcommand\tabularxcolumn[1]{m{#1}}
\begin{tabularx}{\columnwidth}{@{}l*{6}{>{\centering\arraybackslash}X}@{}}
\toprule
Dataset & Nodes &  Node Types & Edges & Edge Types & Target & Classes \\
\midrule
DBLP    & 26,128 & 4 & 239,566 & 6 & author & 4 \\
IMDB    & 21,420 & 4 & 86,642 & 6 & movie & 5 \\
ACM     & 10,942 & 4 & 547,872 & 8 & paper & 3 \\
Freebase& 43,854 & 4 & 151,034 & 6 & movie & 3 \\
\bottomrule
\end{tabularx}
\label{data}
\end{table}

\subsection{Training Objective}
After obtaining the He-Retention output from the final layer, we need a readout function to obtain the embedding of the target node:
\begin{equation}
    \mathbf{H} = \mathrm{Readout}(\mathbf{H}^{S(l)}).
\end{equation}
After obtaining the final representation, we apply it to the specific task and design different loss functions. For supervised node classification, we first map the node representations to the class space through a linear transformation to predict the class distribution, as shown below:
\begin{equation}
    \tilde{\mathbf{Y}} = f_{\Psi}(\mathbf{H}),
\end{equation}
where $\tilde{\mathbf{Y}}\in\mathbb{R}^{N\times C}$ is the prediction and $C$ is the number of classes. To stabilize the optimization process, we further apply L2 regularization on \( \tilde{\mathbf{Y}} \). During training, we apply the cross-entropy loss function to optimize the model:
\begin{equation}
    \mathcal{L} = \sum_{i \in \mathcal{V}_{train}}\mathrm{CrossEntropy}(\tilde{\mathbf{Y}}, \mathbf{Y}_G),
\end{equation}
where $\mathcal{V}_{train}$ is the train set, and $\mathbf{Y}_G$ is ground-truth label.

% To demonstrate the scalability of HeSRN and further reduce complexity for easier inference in node classification tasks, we extend He-Retention into a recurrent form. The recurrent form of He-Retention can be expressed as follows:
% \begin{gather}
% s_{n}=\gamma s_{n-1}+(\mathbf{K}_{n}^{{S}})^\top\mathbf{V}^{S}_{n}+\beta\cdot(\mathbf{K}_{n}^{\mathcal{R}})^{\top}\mathbf{V}^{S}_{n},\\
% \mathrm{He\text{-}Retention}(\mathbf{H}^{S}_n,\mathbf{H}_n^{\mathcal{T}})=\mathbf{Q}^{S}_{n}s_{n}+\beta\cdot\mathbf{Q}_{n}^{\mathcal{R}}s_{n},\quad n=1,...,l
% \label{herr}
% \end{gather}
% where $\mathbf{Q}^{S}, \mathbf{K}^{{S}}, \mathbf{V}^{S}, \mathbf{Q}^{\mathcal{R}}, \mathbf{K}^{\mathcal{R}}, \beta$ are the same as in Equation~\ref{herp}.

\subsection{Complexity Analysis}
In terms of computational complexity, our proposed HeSRN differs significantly from previous model. Since HeSRN adopts slot partitioning with lightweight fusion at the input stage and employs a parallel retentive network for sequence modeling, its main cost comes from node-level slot learning ($\mathcal{O}(N)$) and structure aggregation ($\mathcal{O}(E)$). The heterogeneous graph retentive network cost grows linearly with $\mathcal{O}(N)$, leading to an overall complexity of ($\mathcal{O}(E+N)$), with memory consumption also at ($\mathcal{O}(N)$). In contrast, HINormer, in addition to graph structure and relation encoding, requires performing global self-attention on the context sequence of each target node, where the attention component incurs a cost of ($\mathcal{O}(N L^{2})$), with $L$ denoting the fixed sequence length obtained via sampling. Thus, the overall complexity of HINormer can be expressed as ($\mathcal{O}(E+N L^{2})$), and its memory usage also needs to maintain an ($\mathcal{O}(N L^{2})$) attention map. HeSRN avoids the quadratic attention term and therefore enjoys significantly lower constant factors, making it more efficient for large-scale heterogeneous graphs.
% We have presented two forms of heterogeneous retention above. We refer to the form in Equation~\ref{herp} as the parallel form and the form in Equation~\ref{herr} as the recurrent form. The parallel form allows for the processing of the entire node sequence at once, showcasing powerful parallel computation capabilities. It requires access to the intermediate states of the entire node sequence, resulting in a time complexity of \( O(N^2) \). The recurrent form, on the other hand, relies on local state updates, where the time complexity for each state update is \( O(1) \), making the overall time complexity for the entire node sequence \( O(N) \). Since graph transformer requires a self-attention mechanism to capture dependencies in the node sequence, its time complexity is \( O(N^2) \). Compared to the graph transformer, our HeSRN offers a significant advantage in terms of time complexity. In terms of spatial complexity, graph transformer needs to continuously maintain the Key-Value cache, which has a space complexity of \( O(N) \). Similarly, in HeSRN, the parallel form requires storing retention, and its space complexity is similar to that of graph transformer, \( O(N) \). However, in the recurrent form, only the current state and a small amount of historical information need to be stored, resulting in a space complexity of \( O(1) \).
\begin{table*}[ht]
\centering
\caption{Performance evaluation (\%). OOM mean that the model runs out of memory on the corresponding dataset.}
\resizebox{\linewidth}{!}{
\begin{tabular}{l*{8}{c}} 
\toprule
\multirow{2}{*}{\textbf{Methods}} & \multicolumn{2}{c}{DBLP} & \multicolumn{2}{c}{IMDB} & \multicolumn{2}{c}{ACM} & \multicolumn{2}{c}{Freebase} \\
\cmidrule(lr){2-3} \cmidrule(lr){4-5} \cmidrule(lr){6-7} \cmidrule(l){8-9}
& Micro-F1 & Macro-F1 & Micro-F1 & Macro-F1 & Micro-F1 & Macro-F1 & Micro-F1 & Macro-F1 \\
\midrule
RGCN & 92.07$\pm$0.50 & 91.52$\pm$0.50 & 62.05$\pm$0.15 & 58.85$\pm$0.26 & 91.41$\pm$0.75 & 91.55$\pm$0.74 & 60.82$\pm$1.23 & 59.08$\pm$1.44 \\
HAN & 92.05$\pm$0.62 & 91.67$\pm$0.49 & 64.63$\pm$0.58 & 57.74$\pm$0.96 & 90.79$\pm$0.43 & 90.89$\pm$0.43 & 61.42$\pm$3.56 & 57.05$\pm$2.06 \\
GTN & 93.97$\pm$0.54 & 93.52$\pm$0.55 & 65.14$\pm$0.45 & 60.47$\pm$0.98 & 91.20$\pm$0.71 & 91.31$\pm$0.70 & OOM & OOM \\
HetGNN & 92.33$\pm$0.41 & 91.76$\pm$0.43 & 51.16$\pm$0.65 & 48.25$\pm$0.67 & 86.05$\pm$0.25 & 85.91$\pm$0.25 & 62.99$\pm$2.31 & 58.44$\pm$1.99 \\
MAGNN & 93.76$\pm$0.45 & 93.28$\pm$0.51 & 64.67$\pm$1.67 & 56.49$\pm$3.20 & 90.77$\pm$0.65 & 90.88$\pm$0.64 & 64.43$\pm$0.73 & 58.18$\pm$3.87 \\
HGT & 93.49$\pm$0.25 & 93.01$\pm$0.23 & 67.20$\pm$0.57 & 63.00$\pm$1.19 & 91.00$\pm$0.76 & 91.12$\pm$0.76 & 66.43$\pm$1.88 & 60.03$\pm$2.21 \\
Simple-HGN & 94.46$\pm$0.22 & 94.01$\pm$0.24 & 67.36$\pm$0.57 & 63.53$\pm$1.36 & 93.35$\pm$0.45 & 93.42$\pm$0.44 & 67.49$\pm$0.97 & 62.49$\pm$1.69 \\
\midrule
ANS-GT & 93.15$\pm$0.51 & 92.75$\pm$0.43 & 66.65$\pm$0.35 & 62.52$\pm$0.61 & 92.55$\pm$0.54 & 93.67$\pm$0.62 & 67.33$\pm$0.61 & 61.24$\pm$0.57 \\
NodeFormer & 93.68$\pm$0.42 & 93.05$\pm$0.38 & 65.86$\pm$0.42 & 62.15$\pm$0.77 & 91.89$\pm$0.31 & 92.72$\pm$0.84 & 67.01$\pm$0.52 & 60.83$\pm$1.41 \\
HINormer & 94.94$\pm$0.21 & 94.57$\pm$0.23 & 67.83$\pm$0.34 & 64.65$\pm$0.53 & 93.15$\pm$0.36 & 93.28$\pm$0.43 & 67.78$\pm$0.39 & 62.76$\pm$1.10 \\
PHGT & 95.33$\pm$0.18 & \textbf{94.96}$\pm$0.17 & 68.81$\pm$0.08 & 65.91$\pm$0.30 & 93.72$\pm$0.40 & 93.79$\pm$0.39 & 68.74$\pm$1.42 & 61.73$\pm$1.86 \\
\midrule
HeSRN & \textbf{95.47}$\pm$0.11 & 94.88$\pm$0.12 & \textbf{69.86}$\pm$0.12 & \textbf{65.96}$\pm$0.25 & \textbf{94.25}$\pm$0.20 & \textbf{94.41}$\pm$0.19 & \textbf{69.04}$\pm$0.31 & \textbf{63.25}$\pm$0.67\\
\bottomrule
\end{tabular}}
\label{node}
\end{table*}

\begin{figure}[!ht]
  \centering
  \includegraphics[width=\columnwidth]{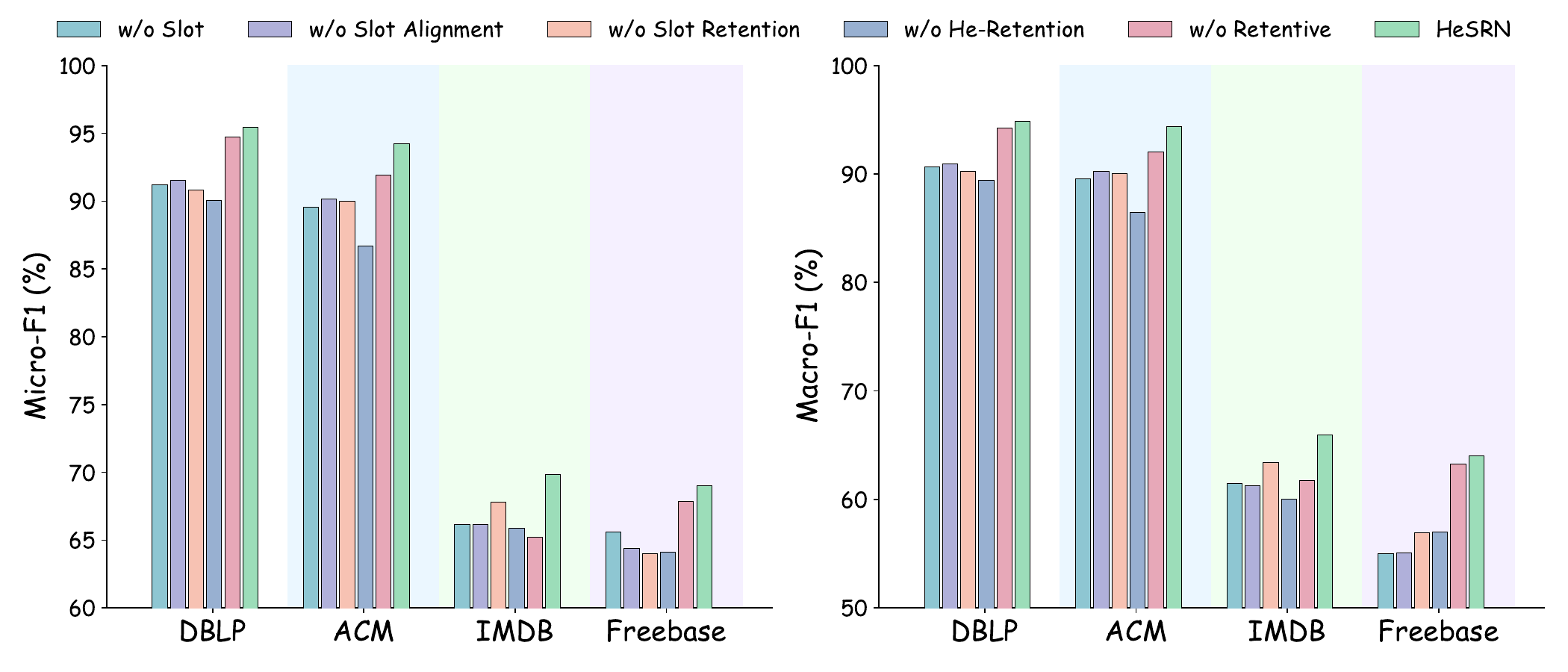}
  \caption{Ablation studies across four datasets.}
\label{abla}
\end{figure}

\begin{figure}[!ht]
  \centering
  \includegraphics[width=\textwidth]{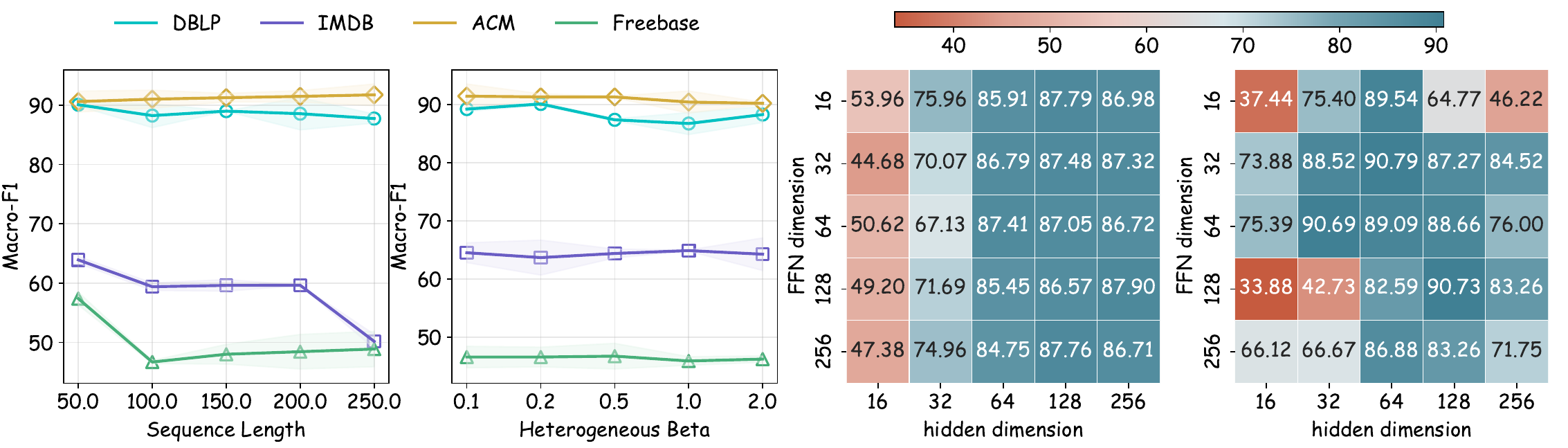}
  \caption{Parameter Sensitivity of Macro-F1 Score.}
\label{fig:sensitivity}
\end{figure}

\section{Experiments}
In this section, we conduct extensive experiments on HeSRN for node classification tasks to evaluate the performance of the proposed model. Additionally, we provide a detailed analysis of the model from several aspects.

\subsection{Experimental Setup}
\subsubsection{Datasets}
We perform our experiments on four commonly utilized heterogeneous graph datasets: DBLP, ACM (both academic graphs), IMDB (a movie graph), and Freebase (a knowledge graph). Table~\ref{data} presents a summary of the statistical details for each dataset. DBLP, ACM, and IMDB are sourced from the Heterogeneous Graph Benchmark (HGB)~\cite{lv2021we}. The data preprocessing and splitting methods adhere to the protocols outlined by HGB. As for Freebase, we utilized the version made available by the latest research~\cite{mao2023hinormer}.

\subsubsection{Baselines}
To comprehensively evaluate our proposed HeSRN against state-of-the-art methods, we consider two categories of baselines: Heterogeneous Graph Neural Networks, including RGCN~\cite{schlichtkrull2018modeling}, HAN~\cite{wang2019heterogeneous}, GTN~\cite{yun2019graph}, HetGNN~\cite{zhang2019heterogeneous}, HGT~\cite{hu2020heterogeneous}, MAGNN~\cite{fu2020magnn}, and Simple-HGN~\cite{lv2021we}, and Transformer-based methods, including ANS-GT~\cite{zhang2022hierarchical}, NodeFormer~\cite{wu2022nodeformer}, HINormer~\cite{mao2023hinormer}, and PHGT~\cite{lu2024heterogeneous}. For the DBLP dataset, we select {APA, APT, APC} as the meta-paths. For the IMDB dataset, we choose {MDM, MAM, MKM} as the meta-paths. For the ACM dataset, we adopt {PAP, PSP} as the meta-paths. For the Freebase dataset, we utilize {MWM, MAM, MDM} as the meta-paths.

\subsubsection{Experimental Settings}
We conducted multi-class node classification experiments on the DBLP, Freebase, and ACM datasets, and multi-label node classification experiments on the IMDB dataset. For each dataset, we randomly partitioned the target nodes into training, validation, and test sets in a ratio of 24:6:70, following the methodology of HGB~\cite{lv2021we}. The performance of node classification is evaluated using the micro-F1 and macro-F1 metrics. All experiments were repeated five times, and we report the averaged results along with their standard deviations. For the baseline methods, we performed hyperparameter tuning based on their official implementations and the recommended settings in HGB. For HeSRN, we explored the following hyperparameter configurations: the hidden layer dimension is set to 256, the learning rate is set to 0.0001, the number of heads is selected from \{2, 4, 6\}, the number of layers in the heterogeneous structure encoder is selected from \{2, 3, 4, 5\}, the number of layers in the Retentive Network is selected from \{2, 3, 4, 5\}, the sequence length is set within the range [20, 200], the heterogeneous relation coefficient $\beta$ is selected from \{0.1, 0.2, 0.5, 1, 2\}, and the dimension of the FFN is selected from \{16, 32, 64, 128\}.

\subsection{Overall Results}
Table~\ref{node} presents the performance of our model and baselines on node classification tasks. It can be observed that our model achieves competitive performance, outperforming other baseline models in most cases. Compared to Transformer-based models, our model performs best in the majority of cases, highlighting the strong potential of the retention mechanism we designed in graph representation learning. HeSRN has a time complexity of \( O(E+N) \) and reduces memory overhead, making it more efficient than Transformer-based models. Notably, on the DBLP dataset, HeSRN's macro-F1 score is second only to the best baseline model, PHGT, but its micro-F1 surpasses all baseline models. We hypothesize that this may be because DBLP data contains more fine-grained information, and PHGT benefits from concatenating cluster tokens, giving it an advantage in capturing fine-grained semantics. We observe that HeSRN outperforms all HGNN baseline models, showcasing its advantages in sequence modeling. Our model can capture long-range dependencies between nodes, naturally avoiding the shortcomings present in HGNNs. Compared to homogeneous Transformer models like ANS-GT and NodeFormer, our model improves micro-F1 by 1.83\%-6.07\% and macro-F1 by 0.79\%-6.13\%, demonstrating the significant advantage of our designed Heterogeneous Retention Encoder. At the same time, our approach surpasses methods that rely on meta-paths (e.g., HAN, MAGNN, and PHGT) in many cases, indicating that models without prior knowledge can still exhibit strong competitiveness.

% \vspace{-10pt}

\subsection{Ablation Studies}
To investigate the contribution of each component in HeSRN, we conduct an ablation study on four datasets as shown in Figure~\ref{abla}. Specifically, w/o Slot replaces the Slot-Aware Structure Encoding module with the initial feature mapping to directly concatenate node representations, w/o Slot Alignment removes the slot alignment operation described in Eq.~\ref{slotalign}, w/o Slot Retention substitutes the retention mechanism with a standard self-attention layer, w/o He-Retention completely discards the heterogeneous input of the retentive architecture by setting $\beta^{\mathcal{T}}=0$, and w/o Retentive replaces the Retentive layer with a Transformer layer. The results demonstrate that removing any of the proposed modules leads to a noticeable performance drop, confirming the need for slot-aware and retentive designs. Specifically, removing the slot construction or the slot alignment module significantly reduces F1 scores, indicating that explicitly separating and aligning type-specific representations is crucial for alleviating semantic entanglement among heterogeneous node types. When the slot retention mechanism is replaced with standard self-attention, the performance degrades further, especially on IMDB and Freebase, which contain richer heterogeneous semantics. This highlights that the exponential-decay retention kernel effectively captures cross-slot correlations in a more stable and efficient way. Similarly, removing the heterogeneous retention encoder notably impairs performance, verifying that the retentive architecture better models long-range dependencies with linear complexity. Overall, the full HeSRN achieves the best results on all datasets, demonstrating that both the slot-aware semantic modeling and the heterogeneous retentive architecture are indispensable for effective and scalable representation learning on heterogeneous graphs.

\subsection{Parameters Sensitivity}
We investigate the sensitivity of HeSRN to key hyperparameters: node sequence length $L$, the heterogeneous retention weight $\beta^{\mathcal{T}}$, and the dimensional relationship between the hidden layer and the feed-forward network (FFN). As shown in Figure~\ref{fig:sensitivity}, the model remains generally stable across a wide range of settings. Performance remains stable within the 50-200 range of sequence length $L$, indicating that the retention mechanism can effectively capture contextual dependencies. When the sequence length increases to 250, performance degradation is observed on IMDB, which may be due to the noise information brought by long-range nodes. The model performs best when $\beta^{\mathcal{T}}$ lies between 0.1 and 0.2, while smaller or larger values lead to minor degradation. This demonstrates that maintaining a balance between type-specific information and global context is crucial for heterogeneous semantic fusion. Regarding the hidden and FFN dimensions, performance improves as the hidden dimension increases from 16 to 128, then saturates or slightly declines at higher dimensions; the FFN dimension exhibits a similar trend, where a moderate increase enhances performance but excessively large values introduce redundancy and slight overfitting. Overall, HeSRN exhibits robust performance across a wide range of hyperparameter settings, demonstrating its stability and scalability.
% We analyze the sensitivity of HeSRN to three key hyperparameters: sequence length $L$, heterogeneous retention weight $\beta^{\mathcal{T}}$, and the dimensional configuration between the hidden layer and FFN. As shown in Figure~\ref{fig:sensitivity}, the model performance remains stable when $L$ ranges from 50 to 200, indicating that HeSRN effectively captures contextual dependencies without requiring long sequences. When the sequence length increases to 250, performance degradation is observed on IMDB, which may be due to the noise information brought by long-order nodes. For $\beta^{\mathcal{T}}$, the best results are obtained when it lies between 0.1 and 0.2, suggesting that a balanced contribution between heterogeneous and global retention is beneficial. Regarding the hidden and FFN dimensions, performance improves as dimensions increase up to 128 but slightly drops beyond that due to redundancy. Overall, HeSRN exhibits robust performance across a wide range of hyperparameter settings, demonstrating its stability and scalability.

% \subsection{Ablation Studies}
% \label{abla}

% \subsection{Parameters Sensitivity}

\section{Conclusion}
In this paper, we proposed HeSRN, a novel Heterogeneous Slot-aware Retentive Network for graph representation learning. HeSRN introduces a slot-aware structure encoder to explicitly disentangle heterogeneous node semantics and a heterogeneous retention encoder to jointly capture both local structural information and global semantic dependencies within node sequences. Compared with Transformer-based models, the retentive architecture significantly reduces computational complexity while maintaining strong expressive capability. By integrating the structural and semantic perspectives of heterogeneous graphs through retention-based fusion, HeSRN provides a unified and efficient framework for heterogeneous graph representation learning. Extensive experiments on four real-world heterogeneous graph datasets demonstrate that HeSRN consistently outperforms representative HGNNs and Graph Transformer baselines on node classification tasks. In future work, we plan to extend HeSRN to heterogeneous dynamic graphs by designing temporal retention mechanisms that can dynamically capture evolving heterogeneous semantics over time.
\clearpage
\bibliography{main}

@inproceedings{chenleveraging,
  title={Leveraging Contrastive Learning for Enhanced Node Representations in Tokenized Graph Transformers},
  author={Chen, Jinsong and Liu, Hanpeng and Hopcroft, John E and He, Kun},
  year={2024},
  booktitle={The Thirty-eighth Annual Conference on Neural Information Processing Systems}
}

@inproceedings{shomer2024lpformer,
  title={Lpformer: An adaptive graph transformer for link prediction},
  author={Shomer, Harry and Ma, Yao and Mao, Haitao and Li, Juanhui and Wu, Bo and Tang, Jiliang},
  booktitle={Proceedings of the 30th ACM SIGKDD Conference on Knowledge Discovery and Data Mining},
  pages={2686--2698},
  year={2024}
}

@inproceedings{chen2024sigformer,
  title={SIGformer: Sign-aware Graph Transformer for Recommendation},
  author={Chen, Sirui and Chen, Jiawei and Zhou, Sheng and Wang, Bohao and Han, Shen and Su, Chanfei and Yuan, Yuqing and Wang, Can},
  booktitle={Proceedings of the 47th International ACM SIGIR Conference on Research and Development in Information Retrieval},
  pages={1274--1284},
  year={2024}
}

@article{zou2025loha,
  title={LOHA: Direct Graph Spectral Contrastive Learning Between Low-pass and High-pass Views},
  author={Zou, Ziyun and Jiang, Yinghui and Shen, Lian and Liu, Juan and Liu, Xiangrong},
  journal={arXiv preprint arXiv:2501.02969},
  year={2025}
}

@article{luan2024graph,
  title={When do graph neural networks help with node classification? investigating the homophily principle on node distinguishability},
  author={Luan, Sitao and Hua, Chenqing and Xu, Minkai and Lu, Qincheng and Zhu, Jiaqi and Chang, Xiao-Wen and Fu, Jie and Leskovec, Jure and Precup, Doina},
  journal={Advances in Neural Information Processing Systems},
  volume={36},
  year={2024}
}

@inproceedings{das2024ags,
  title={Ags-gnn: Attribute-guided sampling for graph neural networks},
  author={Das, Siddhartha Shankar and Ferdous, SM and Halappanavar, Mahantesh M and Serra, Edoardo and Pothen, Alex},
  booktitle={Proceedings of the 30th ACM SIGKDD Conference on Knowledge Discovery and Data Mining},
  pages={538--549},
  year={2024}
}

@inproceedings{brodyattentive,
  title={How Attentive are Graph Attention Networks?},
  author={Brody, Shaked and Alon, Uri and Yahav, Eran},
  booktitle={International Conference on Learning Representations},
  year={2021}
}

@inproceedings{velivckovic2018graph,
  title={Graph Attention Networks},
  author={Veli{\v{c}}kovi{\'c}, Petar and Cucurull, Guillem and Casanova, Arantxa and Romero, Adriana and Li{\`o}, Pietro and Bengio, Yoshua},
  booktitle={International Conference on Learning Representations},
  year={2018}
}

@inproceedings{ji2024regcl,
  title={ReGCL: Rethinking Message Passing in Graph Contrastive Learning},
  author={Ji, Cheng and Huang, Zixuan and Sun, Qingyun and Peng, Hao and Fu, Xingcheng and Li, Qian and Li, Jianxin},
  booktitle={Proceedings of the AAAI Conference on Artificial Intelligence},
  volume={38},
  number={8},
  pages={8544--8552},
  year={2024}
}

@article{wang2024graph,
  title={Graph-mamba: Towards long-range graph sequence modeling with selective state spaces},
  author={Wang, Chloe and Tsepa, Oleksii and Ma, Jun and Wang, Bo},
  journal={arXiv preprint arXiv:2402.00789},
  year={2024}
}

@inproceedings{shen2024resisting,
  title={Resisting over-smoothing in graph neural networks via dual-dimensional decoupling},
  author={Shen, Wei and Ye, Mang and Huang, Wenke},
  booktitle={Proceedings of the 32nd ACM International Conference on Multimedia},
  pages={5800--5809},
  year={2024}
}

@inproceedings{huanguniversal,
  title={How Universal Polynomial Bases Enhance Spectral Graph Neural Networks: Heterophily, Over-smoothing, and Over-squashing},
  author={Huang, Keke and Wang, Yu Guang and Li, Ming and Lio, Pietro},
  booktitle={Forty-first International Conference on Machine Learning},
  year={2024}
}

@inproceedings{zhu2023structural,
  title={On structural expressive power of graph transformers},
  author={Zhu, Wenhao and Wen, Tianyu and Song, Guojie and Wang, Liang and Zheng, Bo},
  booktitle={Proceedings of the 29th ACM SIGKDD Conference on Knowledge Discovery and Data Mining},
  pages={3628--3637},
  year={2023}
}

@article{lu2023neighborhood,
  title={Neighborhood overlap-aware heterogeneous hypergraph neural network for link prediction},
  author={Lu, Yifan and Gao, Mengzhou and Liu, Huan and Liu, Zehao and Yu, Wei and Li, Xiaoming and Jiao, Pengfei},
  journal={Pattern Recognition},
  volume={144},
  pages={109818},
  year={2023},
  publisher={Elsevier}
}

@inproceedings{boker2024fine,
  title={Fine-grained expressivity of graph neural networks},
  author={B{\"o}ker, Jan and Levie, Ron and Huang, Ningyuan and Villar, Soledad and Morris, Christopher},
  booktitle={Proceedings of the 37th International Conference on Neural Information Processing Systems},
  pages={46658--46700},
  year={2023}
}

@article{retnet,
  title={Retentive network: A successor to transformer for large language models},
  author={Sun, Yutao and Dong, Li and Huang, Shaohan and Ma, Shuming and Xia, Yuqing and Xue, Jilong and Wang, Jianyong and Wei, Furu},
  journal={arXiv preprint arXiv:2307.08621},
  year={2023}
}

@inproceedings{huang2024leret,
  title={Leret: Language-empowered retentive network for time series forecasting},
  author={Huang, Qihe and Zhou, Zhengyang and Yang, Kuo and Lin, Gengyu and Yi, Zhongchao and Wang, Yang},
  booktitle={Proceedings of the Thirty-Third International Joint Conference on Artificial Intelligence, IJCAI-24},
  year={2024}
}

@article{guo2024vl,
  title={VL-MFER: A Vision-Language Multimodal Pretrained Model with Multiway-Fuzzy-Experts Bidirectional Retention Network},
  author={Guo, Chen and Li, Xinran and Ma, Jiaman and Li, Yimeng and Liu, Yuefan and Qi, Haiying and Zhang, Li and Jin, Yuhan},
  journal={IEEE Transactions on Fuzzy Systems},
  year={2024},
  publisher={IEEE}
}

@article{zaremba2014recurrent,
  title={Recurrent neural network regularization},
  author={Zaremba, Wojciech},
  journal={arXiv preprint arXiv:1409.2329},
  year={2014}
}

@article{vaswani2017attention,
  title={Attention is all you need},
  author={Vaswani, A},
  journal={Advances in Neural Information Processing Systems},
  year={2017}
}

@article{kazemnejad2024impact,
  title={The impact of positional encoding on length generalization in transformers},
  author={Kazemnejad, Amirhossein and Padhi, Inkit and Natesan Ramamurthy, Karthikeyan and Das, Payel and Reddy, Siva},
  journal={Advances in Neural Information Processing Systems},
  volume={36},
  year={2024}
}

@inproceedings{ruoss2023randomized,
  title={Randomized Positional Encodings Boost Length Generalization of Transformers},
  author={Ruoss, Anian and Del{\'e}tang, Gr{\'e}goire and Genewein, Tim and Grau-Moya, Jordi and Csord{\'a}s, R{\'o}bert and Bennani, Mehdi and Legg, Shane and Veness, Joel},
  booktitle={Proceedings of the 61st Annual Meeting of the Association for Computational Linguistics (Volume 2: Short Papers)},
  pages={1889--1903},
  year={2023}
}

@inproceedings{wu2021rethinking,
  title={Rethinking and improving relative position encoding for vision transformer},
  author={Wu, Kan and Peng, Houwen and Chen, Minghao and Fu, Jianlong and Chao, Hongyang},
  booktitle={Proceedings of the IEEE/CVF International Conference on Computer Vision},
  pages={10033--10041},
  year={2021}
}

@inproceedings{sun2023length,
  title={A Length-Extrapolatable Transformer},
  author={Sun, Yutao and Dong, Li and Patra, Barun and Ma, Shuming and Huang, Shaohan and Benhaim, Alon and Chaudhary, Vishrav and Song, Xia and Wei, Furu},
  booktitle={Proceedings of the 61st Annual Meeting of the Association for Computational Linguistics (Volume 1: Long Papers)},
  pages={14590--14604},
  year={2023}
}

@article{jiang2023pharmacophoric,
  title={Pharmacophoric-constrained heterogeneous graph transformer model for molecular property prediction},
  author={Jiang, Yinghui and Jin, Shuting and Jin, Xurui and Xiao, Xianglu and Wu, Wenfan and Liu, Xiangrong and Zhang, Qiang and Zeng, Xiangxiang and Yang, Guang and Niu, Zhangming},
  journal={Communications Chemistry},
  volume={6},
  number={1},
  pages={60},
  year={2023},
  publisher={Nature Publishing Group UK London}
}

@inproceedings{lu2024heterogeneous,
  title={Heterogeneous graph transformer with poly-tokenization},
  author={Lu, Zhiyuan and Fang, Yuan and Yang, Cheng and Shi, Chuan},
  year={2024},
  booktitle={International Joint Conferences on Artificial Intelligence}
}

@article{kipf2016semi,
  title={Semi-supervised classification with graph convolutional networks},
  author={Kipf, Thomas N and Welling, Max},
  journal={arXiv preprint arXiv:1609.02907},
  year={2016}
}

@inproceedings{wu2018group,
  title={Group normalization},
  author={Wu, Yuxin and He, Kaiming},
  booktitle={Proceedings of the European conference on computer vision (ECCV)},
  pages={3--19},
  year={2018}
}

@article{hendrycks2016gaussian,
  title={Gaussian error linear units (gelus)},
  author={Hendrycks, Dan and Gimpel, Kevin},
  journal={arXiv preprint arXiv:1606.08415},
  year={2016}
}

@article{lei2016layer,
  title={Layer normalization},
  author={Lei Ba, Jimmy and Kiros, Jamie Ryan and Hinton, Geoffrey E},
  journal={ArXiv e-prints},
  pages={arXiv--1607},
  year={2016}
}

@inproceedings{lv2021we,
  title={Are we really making much progress? revisiting, benchmarking and refining heterogeneous graph neural networks},
  author={Lv, Qingsong and Ding, Ming and Liu, Qiang and Chen, Yuxiang and Feng, Wenzheng and He, Siming and Zhou, Chang and Jiang, Jianguo and Dong, Yuxiao and Tang, Jie},
  booktitle={Proceedings of the 27th ACM SIGKDD conference on knowledge discovery \& data mining},
  pages={1150--1160},
  year={2021}
}

@inproceedings{mao2023hinormer,
  title={Hinormer: Representation learning on heterogeneous information networks with graph transformer},
  author={Mao, Qiheng and Liu, Zemin and Liu, Chenghao and Sun, Jianling},
  booktitle={Proceedings of the ACM Web Conference 2023},
  pages={599--610},
  year={2023}
}

@inproceedings{schlichtkrull2018modeling,
  title={Modeling relational data with graph convolutional networks},
  author={Schlichtkrull, Michael and Kipf, Thomas N and Bloem, Peter and Van Den Berg, Rianne and Titov, Ivan and Welling, Max},
  booktitle={The semantic web: 15th international conference, ESWC 2018, Heraklion, Crete, Greece, June 3--7, 2018, proceedings 15},
  pages={593--607},
  year={2018},
  organization={Springer}
}

@inproceedings{wang2019heterogeneous,
  title={Heterogeneous graph attention network},
  author={Wang, Xiao and Ji, Houye and Shi, Chuan and Wang, Bai and Ye, Yanfang and Cui, Peng and Yu, Philip S},
  booktitle={The world wide web conference},
  pages={2022--2032},
  year={2019}
}

@article{yun2019graph,
  title={Graph transformer networks},
  author={Yun, Seongjun and Jeong, Minbyul and Kim, Raehyun and Kang, Jaewoo and Kim, Hyunwoo J},
  journal={Advances in neural information processing systems},
  volume={32},
  year={2019}
}

@inproceedings{zhang2019heterogeneous,
  title={Heterogeneous graph neural network},
  author={Zhang, Chuxu and Song, Dongjin and Huang, Chao and Swami, Ananthram and Chawla, Nitesh V},
  booktitle={Proceedings of the 25th ACM SIGKDD international conference on knowledge discovery \& data mining},
  pages={793--803},
  year={2019}
}

@inproceedings{hu2020heterogeneous,
  title={Heterogeneous graph transformer},
  author={Hu, Ziniu and Dong, Yuxiao and Wang, Kuansan and Sun, Yizhou},
  booktitle={Proceedings of the web conference 2020},
  pages={2704--2710},
  year={2020}
}

@inproceedings{fu2020magnn,
  title={Magnn: Metapath aggregated graph neural network for heterogeneous graph embedding},
  author={Fu, Xinyu and Zhang, Jiani and Meng, Ziqiao and King, Irwin},
  booktitle={Proceedings of the web conference 2020},
  pages={2331--2341},
  year={2020}
}

@article{zhang2022hierarchical,
  title={Hierarchical graph transformer with adaptive node sampling},
  author={Zhang, Zaixi and Liu, Qi and Hu, Qingyong and Lee, Chee-Kong},
  journal={Advances in Neural Information Processing Systems},
  volume={35},
  pages={21171--21183},
  year={2022}
}

@article{wu2022nodeformer,
  title={Nodeformer: A scalable graph structure learning transformer for node classification},
  author={Wu, Qitian and Zhao, Wentao and Li, Zenan and Wipf, David P and Yan, Junchi},
  journal={Advances in Neural Information Processing Systems},
  volume={35},
  pages={27387--27401},
  year={2022}
}

@article{rampavsek2022recipe,
  title={Recipe for a general, powerful, scalable graph transformer},
  author={Ramp{\'a}{\v{s}}ek, Ladislav and Galkin, Michael and Dwivedi, Vijay Prakash and Luu, Anh Tuan and Wolf, Guy and Beaini, Dominique},
  journal={Advances in Neural Information Processing Systems},
  volume={35},
  pages={14501--14515},
  year={2022}
}

@article{wu2024simplifying,
  title={Simplifying and empowering transformers for large-graph representations},
  author={Wu, Qitian and Zhao, Wentao and Yang, Chenxiao and Zhang, Hengrui and Nie, Fan and Jiang, Haitian and Bian, Yatao and Yan, Junchi},
  journal={Advances in Neural Information Processing Systems},
  volume={36},
  year={2024}
}

@inproceedings{xingless,
  title={Less is More: on the Over-Globalizing Problem in Graph Transformers},
  author={Xing, Yujie and Wang, Xiao and Li, Yibo and Huang, Hai and Shi, Chuan},
  booktitle={Forty-first International Conference on Machine Learning},
  year={2024}
}

@inproceedings{kong2023goat,
  title={GOAT: A global transformer on large-scale graphs},
  author={Kong, Kezhi and Chen, Jiuhai and Kirchenbauer, John and Ni, Renkun and Bruss, C Bayan and Goldstein, Tom},
  booktitle={International Conference on Machine Learning},
  pages={17375--17390},
  year={2023},
  organization={PMLR}
}

@inproceedings{zhang2024cached,
  title={Cached Transformers: Improving Transformers with Differentiable Memory Cachde},
  author={Zhang, Zhaoyang and Shao, Wenqi and Ge, Yixiao and Wang, Xiaogang and Gu, Jinwei and Luo, Ping},
  booktitle={Proceedings of the AAAI Conference on Artificial Intelligence},
  volume={38},
  number={15},
  pages={16935--16943},
  year={2024}
}

@article{brandon2024reducing,
  title={Reducing Transformer Key-Value Cache Size with Cross-Layer Attention},
  author={Brandon, William and Mishra, Mayank and Nrusimha, Aniruddha and Panda, Rameswar and Kelly, Jonathan Ragan},
  journal={arXiv preprint arXiv:2405.12981},
  year={2024}
}

@inproceedings{yang2023simple,
  title={Simple and efficient heterogeneous graph neural network},
  author={Yang, Xiaocheng and Yan, Mingyu and Pan, Shirui and Ye, Xiaochun and Fan, Dongrui},
  booktitle={Proceedings of the AAAI conference on artificial intelligence},
  volume={37},
  number={9},
  pages={10816--10824},
  year={2023}
}

@article{wang2022survey,
  title={A survey on heterogeneous graph embedding: methods, techniques, applications and sources},
  author={Wang, Xiao and Bo, Deyu and Shi, Chuan and Fan, Shaohua and Ye, Yanfang and Philip, S Yu},
  journal={IEEE Transactions on Big Data},
  volume={9},
  number={2},
  pages={415--436},
  year={2022},
  publisher={IEEE}
}

@inproceedings{wu2023molformer,
  title={Molformer: Motif-based transformer on 3d heterogeneous molecular graphs},
  author={Wu, Fang and Radev, Dragomir and Li, Stan Z},
  booktitle={Proceedings of the AAAI Conference on Artificial Intelligence},
  volume={37},
  number={4},
  pages={5312--5320},
  year={2023}
}

@inproceedings{yang2023revisiting,
  title={Revisiting citation prediction with cluster-aware text-enhanced heterogeneous graph neural networks},
  author={Yang, Carl and Han, Jiawei},
  booktitle={2023 IEEE 39th International Conference on Data Engineering (ICDE)},
  pages={682--695},
  year={2023},
  organization={IEEE}
}

@article{ruiz2024high,
  title={High dimensional, tabular deep learning with an auxiliary knowledge graph},
  author={Ruiz, Camilo and Ren, Hongyu and Huang, Kexin and Leskovec, Jure},
  journal={Advances in Neural Information Processing Systems},
  volume={36},
  year={2024}
}

@inproceedings{he2016deep,
  title={Deep residual learning for image recognition},
  author={He, Kaiming and Zhang, Xiangyu and Ren, Shaoqing and Sun, Jian},
  booktitle={Proceedings of the IEEE conference on computer vision and pattern recognition},
  pages={770--778},
  year={2016}
}

@inproceedings{ni2025robust,
  title={Robust Heterogeneous Graph Classification for Molecular Property Prediction with Information Bottleneck},
  author={Ni, Zhibin and Liu, Chang and Wan, Hai and Zhao, Xibin},
  booktitle={Proceedings of the AAAI Conference on Artificial Intelligence},
  volume={39},
  number={1},
  pages={640--648},
  year={2025}
}

@inproceedings{li2023heterogeneous,
  title={Heterogeneous temporal graph neural network explainer},
  author={Li, Jiazheng and Zhang, Chunhui and Zhang, Chuxu},
  booktitle={Proceedings of the 32nd ACM International Conference on Information and Knowledge Management},
  pages={1298--1307},
  year={2023}
}

@inproceedings{tang2024higpt,
  title={Higpt: Heterogeneous graph language model},
  author={Tang, Jiabin and Yang, Yuhao and Wei, Wei and Shi, Lei and Xia, Long and Yin, Dawei and Huang, Chao},
  booktitle={Proceedings of the 30th ACM SIGKDD conference on knowledge discovery and data mining},
  pages={2842--2853},
  year={2024}
}

@article{chen2022nagphormer,
  title={NAGphormer: A tokenized graph transformer for node classification in large graphs},
  author={Chen, Jinsong and Gao, Kaiyuan and Li, Gaichao and He, Kun},
  journal={arXiv preprint arXiv:2206.04910},
  year={2022}
}

\end{document}